%% file: main.tex
\documentclass[10pt,twocolumn,letterpaper]{article}

\usepackage{wacv}              


\usepackage{rotating} 
\usepackage{multirow}
\usepackage{float}
\usepackage{graphicx}
\usepackage{subcaption}
\usepackage{makecell} 
\usepackage[utf8]{inputenc}
\usepackage{newunicodechar}
\newunicodechar{✘}{\ding{55}} 
\newunicodechar{✔}{\ding{51}}
\usepackage{pifont} 
\definecolor{wacvblue}{rgb}{0.21,0.49,0.74}
\usepackage[pagebackref,breaklinks,colorlinks,allcolors=wacvblue]{hyperref}

\def\wacvPaperID{558} 
\def\confName{WACV}
\def\confYear{2026}

\title{UCDSC: Open Set UnCertainty aware Deep Simplex Classifier for Medical Image Datasets
\thanks{Accepted at IEEE WACV 2026}
}
\author{
Arnav Aditya,
Nitin Kumar, and
Saurabh Shigwan\\
\textit{Shiv Nadar Institution of Eminence, Delhi NCR, India}\\
{\tt\small \{aa716, nitin.kumar, saurabh.shigwan\}@snu.edu.in}
}

\begin{document}
\maketitle
\input{sec/0_abstract}    
\input{sec/1_intro}

\input{sec/2_method_result}

\input{sec/3_conclusion}
{
    \small
\bibliographystyle{ieeenat_fullname}
    \bibliography{main}
}
\end{document}

%% file: sec/0_abstract.tex
\begin{abstract}
Driven by advancements in deep learning, computer-aided diagnoses have made remarkable progress. However, outside controlled laboratory settings,  algorithms may encounter several challenges. In the medical domain, these difficulties often stem from limited data availability due to ethical and legal restrictions, as well as the high cost and time required for expert annotations—especially in the face of emerging or rare diseases. In this context, open-set recognition plays a vital role by identifying whether a sample belongs to one of the known classes seen during training or should be rejected as an unknown. Recent studies have shown that features learned in the later stages of deep neural networks are observed to cluster around their class means, which themselves are arranged as individual vertices of a regular simplex \cite{papyan2020prevalence}. The proposed method introduces a loss function designed to reject samples of unknown classes effectively by penalizing open space regions using auxiliary datasets. This approach achieves significant performance gain across four MedMNIST datasets—BloodMNIST, OCTMNIST, DermaMNIST,  TissueMNIST and a publicly available skin dataset\cite{naqvi2023augmented} outperforming state-of-the-art techniques.
\end{abstract}

%% file: sec/1_intro.tex
\section{Introduction}
\label{sec:intro}


%
Open-Set Recognition (OSR) formalizes the scenario in which a classifier must handle classes unseen during training. 
Although closed set classification assumes a fully known label space shared between training and testing, OSR acknowledges the open world nature of real world data by allowing unknown classes to appear during inference \cite{geng2021recent,sun2023comprehensive}.
In other words, the model must not only correctly classify inputs from \textbf{known} classes (seen during training) but also detect or reject inputs from \textbf{unknown} classes (entirely new) \cite{scheirer2013toward}.
%
At inference, OSR models often assign a special ``unknown'' label or apply confidence thresholds to reject unfamiliar samples \cite{bendale2015towards}.

To enhance the classification performance of deep neural network models, significant research has been directed towards maximizing the inter-class margins. Recent approaches in this area can be broadly categorized into two groups based on whether margin maximization is performed in Euclidean space or in angular space~\cite{arcface, cosface}. Methods operating in Euclidean space aim to reduce intra-class distances while increasing inter-class separations using Euclidean distance metrics, which are typically employed during the inference stage. In contrast, approaches based on angular space rely on cosine similarity, focusing on maximizing angular margins between classes for improved discriminative capability during classification.

In medical imaging, OSR plays a critical role due to the inherently open-set nature of clinical environments, where rare diseases, imaging artifacts, or previously unidentified pathologies frequently arise. Diagnostic deep learning models trained under closed set assumptions can confidently assign incorrect labels to such unfamiliar cases, leading to misdiagnosis and potentially severe clinical consequences. Integrating OSR mechanisms into medical imaging workflows enables these systems to identify out-of-distribution or novel inputs, promoting safer and more trustworthy AI-assisted decision-making.

While neural networks have achieved remarkable success in image classification, significant challenges persist in the medical imaging domain—particularly when dealing with emerging rare, or new diseases. These challenges are exacerbated by the limited availability and incomplete collection of annotated training datasets. Recent theoretical insights reveal that, in the final phase of training, deep neural classifiers tend to exhibit a geometric convergence: the last-layer feature representations collapse to their respective class means, and these class means, along with the corresponding classifier weight vectors, align with the vertices of a simplex equiangular tight frame (ETF). This phenomenon is known as Neural Collapse (NC)~\cite{papyan2020prevalence}.


Our method is based on leveraging the phenomenon of Neural Collapse on simplex ETF in which we jointly optimize the ResNet18/34 as a backbone with the proposed loss function in equation
~\eqref{eq:ltotal}. The contributions of this paper are summarized as follows: 1) Our proposed framework incorporates euclidean distance as well as angular distance through one-to-one mapping of the class centers to the vertices of the regular simplex (Deep Simplex Classifier (DSC)) \cite{cevikalp2024reaching} on medical image datasets.
2) Our proposed loss function includes uncertainty aware regularization term which penalizes the open-space between any two class centers and thereby improving the performance of the rejections of samples belonging to the unknown class(es). 3) Our proposed loss function includes component for auxiliary datasets to push unknown class regions away from the known class centers. 

\subsection*{Theoretical Foundations}

Scheirer et al.~\cite{scheirer2013toward} first formalized OSR by introducing the concept of \textit{open space risk}, defining a classifier $f(x) > 0$ to indicate a known-class decision and minimizing an objective that includes both empirical risk and the risk over open space:
\begin{equation}
    R_{\text{open}}(f) = \lambda R_\varepsilon(f) + R_O(f),
\end{equation}
where $R_\varepsilon$ is the empirical classification loss and $R_O$ measures how much of the open space is incorrectly labeled as known. This formulation encourages classifiers to shrink decision regions around training data to avoid incorrect labeling of unknowns.

To reduce open space risk, Scheirer et al.~\cite{scheirer2014probability} proposed Compact Abating Probability (CAP) models, which force class probabilities to decay as one moves away from known examples, often modeled using extreme value theory (EVT)~\cite{coles2001introduction}. One such example is the Extreme Value Machine (EVM)~\cite{rudd2018extreme}, which fits Weibull distributions to distances in feature space, estimating the tail probabilities to detect novelty.

In summary, OSR extends statistical learning by penalizing decisions in regions unrepresented by training data, bridging classification, and outlier detection.

\section{Related Work}

Open-set recognition (OSR) addresses the challenge of correctly classifying test samples from known classes while rejecting samples from unknown classes not encountered during training. Scheirer et al.~\cite{scheirer2013toward} formalized the concept of OSR, highlighting the limitations of traditional closed-set classifiers.
In general, OSR methods must both classify in-distribution samples correctly and reject or defer out-of-distribution samples. Geng \textit{et al.}~\cite{geng2021recent} define this dual goal succinctly: ``OSR requires the classifiers to not only accurately classify the seen classes, but also effectively deal with unseen ones''.

Subsequently, Bendale and Boult~\cite{bendale2016towards} proposed the OpenMax model, which replaces softmax with a mechanism for detecting unknowns via statistical modeling of activation vectors.
%
A simple baseline is to \textit{threshold the Softmax confidence}: inputs with maximum softmax below a threshold are labeled ``unknown''. However, raw softmax scores are often overconfident on outliers \cite{liu2020energy}. OpenMax\cite{bendale2016towards} and its variant G-OpenMax\cite{BMVC2017_42} replace softmax with an extreme-value-theory (EVT) calibration~\cite{scheirer2022extreme,vignotto2018extreme}. Shu \textit{et al.} 
propose DOC~\cite{shu2017doc}, which deploys $m$ 1-vs-rest sigmoids for $m$ known classes and based on the corresponding probability thresholds it accepts or rejects the samples. 
Reconstruction-based models like CROSR~\cite{yoshihashi2019classification} and C2AE~\cite{oza2019c2ae} learn auxiliary decoders to reconstruct known samples, making reconstruction error a proxy for novelty.

%

Separate from end-to-end DNNs, OSR methods also use distance metrics in feature space to detect novelties. Classic approaches extend nearest-prototype or nearest-neighbor classifiers with rejection rules. Nearest Non-Outlier (NNO) by Bendale \& Boult \cite{bendale2015towards} extends Nearest Class Mean (NCM) classification: an input is assigned the label of the closest class mean if it lies within a class-specific radius, otherwise it is rejected as unknown.
Open-set Nearest-Neighbor (OSNN) \cite{mendes2017nearest} classification improves on standard k-NN by applying a distance ratio test: if the ratio of distances to the nearest and second-nearest neighbors (from different classes) exceeds a threshold, the sample is classified; otherwise it is marked unknown.
This nearest-neighbor distance-ratio (NNDR) rule filters ambiguous cases. A more sophisticated metric approach is the Extreme Value Machine (EVM) \cite{rudd2018extreme}, which models each training sample’s margin distances with a Weibull distribution. At test time, EVM computes the probability that a sample belongs to each known class based on its distance and rejects it if all class probabilities fall below a cutoff.
The phenomenon of neural collapse, where deep features converge around class means lying on a regular simplex~\cite{papyan2020prevalence}, has inspired recent OSR methods that use geometric alignment of embeddings to distinguish unknown classes \cite{zhu2021geometric,thrampoulidis2022imbalance,yaras2022neural,gill2024engineering}.

In medical imaging, OSR techniques have only recently been explored. The MedMNIST benchmark (a collection of standardized 2D and 3D biomedical image datasets) is often used for prototyping OSR ideas~\cite{yang2021medmnist}. For example, Jia \textit{et al.}~\cite{jia2024revealing} evaluate open-set models on subsets of MedMNIST v2 (e.g., ChestMNIST, OCTMNIST, PneumoniaMNIST), showing that a distillation-based network can improve OSR accuracy on these tasks. Wang et al.~\cite{wang2023uncertainty} tackle this by introducing the Uncertainty-Inspired Open-Set (UIOS) framework for retinal disease classification. Their method uses evidential deep networks~\cite{sensoy2018evidential} to estimate both class probabilities and predictive uncertainty, effectively flagging out-of-distribution (OOD) inputs like rare diseases or artifacts.
Despite significant progress, developing robust open-set recognition models under data limitations and inter-class similarities remains an ongoing challenge in medical AI. The study emphasizes that OSR is crucial across medical domains to avoid misdiagnoses from unfamiliar cases and to ensure safe deferral to human experts.

Generative OSR models synthesize pseudo-unknowns or model densities. G-OpenMax~\cite{ge2017generative} and OSRCI~\cite{neal2018open} use GANs to train with synthetic unknowns.  OpenHybrid~\cite{zhang2020hybrid} jointly trains a CNN classifier and a flow-based density estimator.
Advances in prototype-based learning have influenced open-set recognition by promoting structured feature spaces. GCPL (Convolutional Prototype Learning)~\cite{8578464} uses a prototype representation for each known class and applies a prototype loss to encourage embeddings to cluster around their respective centers. This reduces intra-class variance but does not model unknowns explicitly. As a result, unknown samples may occupy the same regions as known classes. Prototypes may also drift into regions containing unknowns early in training, increasing open space risk. 

RPL (Reciprocal Point Learning)~\cite{10.1007/978-3-030-58580-8_30} models the extra-class space using learnable reciprocal points trained via a 1-vs-rest scheme. This improves awareness of open space but enforces fixed-margin constraints and ignores angular relationships, making the method sensitive to margin size and initialization. These limitations are addressed by DSC~\cite{cevikalp2024reaching} which incorporates an intra-class loss that pulls samples of the same class closer to their class center, while class centers are fixed on the vertices of a simplex ETF. This ensures maximal inter-class angular separation without the need for an explicit inter-class term. However, DSC partially addresses the challenge of rejecting an unknown sample by training on an auxiliary dataset of background samples.   

ARPL (Adversarial Reciprocal Point Learning)~\cite{9521769} extends RPL by using an adversarial margin constraint, angle-based similarity, and open-space regularization to prevent overlap between known and unknown regions. It also introduces confusing samples generated through adversarial learning to simulate domain-shifted unknowns and adapts the feature space size to the number of known classes.We evaluated ARPL+CS (Confusing Samples) approach mentioned in~\cite{9521769} for comparison with our proposed method.We avoid synthetic augmentation and instead use an uncertainty loss to recover misclassified unknowns, thus achieving better separation using only learned discriminative signals.

DIAS (Difficulty-Aware Simulator)~\cite{moon2022difficulty} takes a generative approach by simulating unknowns of varying difficulty through a Copycat generator that mimics the classifier. This exposes the classifier to hard-to-distinguish unknowns and helps maintain calibrated decisions. In contrast, our method differs by not relying on generative mechanisms; instead, we integrate intra-class, outlier triplet and uncertainty losses to enforce decision boundary structure and reduce overconfidence in open-set conditions.

OMCL (Open Margin Cosine Loss)~\cite{liu2023omcl} addresses open-set recognition by explicitly leveraging the structure of the embedding space, where known class features are encouraged to cluster compactly and sparse regions are treated as unknowns. It combines Margin Loss with Adaptive Scale (MLAS) and Open-Space Suppression (OSS) to achieve this. MLAS incorporates an angular margin and a learnable scaling factor to reinforce intra-class compactness and inter-class separability. OSS further enhances open-set detection by generating synthetic descriptors that populate sparse areas of the feature space and explicitly categorizing them as unknown.  
However, OMCL is based on the assumption that known classes densely occupy certain parts of feature space, and unknowns lie in sparse regions, it might fail if unknowns are somewhat close (in feature space) to the knowns. Our proposed method addresses this by including the uncertainty aware term which penalizes the open space between the class centers.

Compared to all of the above, our method combines three complementary objectives: intra-class loss, outlier triplet loss, and uncertainty loss~\cite{cao2021open} to simultaneously promote class compactness, maximize separability, improve rejection of difficult unknowns, and minimize open space risk. This integrated discriminative framework offers stable and interpretable performance across various open-set environments without the need for adversarial training, generative modeling, or tuning multiple hyper-parameters. Table \ref{tab:methods} provides information on various state-of-the-art methods using the auxiliary datasets, synthetic samples, and explicitly modeling the open space. 
%

\begin{table}[htbp]
\centering
\caption{Comparison of methods on auxiliary dataset usage, open space modeling, and synthetic sample generation.}
\label{tab:methods}
\resizebox{\columnwidth}{!}{%
\begin{tabular}{lccc}
\toprule
\textbf{Method Name} & \makecell{\textbf{Uses Auxiliary} \\ \textbf{Datasets}} & \makecell{\textbf{Explicitly Models} \\ \textbf{Open Space}} & \makecell{\textbf{Requires Synthetic} \\ \textbf{Samples}} \\
\midrule
GCPL~\cite{8578464}    & ✘ & ✘ & ✘ \\
RPL~\cite{10.1007/978-3-030-58580-8_30}     & ✘ & ✔ & ✘ \\
ARPL+CS~\cite{9521769} & ✘ & ✔ & ✔ \\
DIAS~\cite{moon2022difficulty}    & ✘ & ✔ & ✔ \\
OMCL~\cite{liu2023omcl}    & ✘ & ✔ & ✔ \\
DSC~\cite{cevikalp2024reaching}     & ✔ & ✘ & ✘ \\
UCDSC(\textbf{Ours})   & ✔ & ✔ & ✘ \\
\bottomrule
\end{tabular}%
}
\end{table}

%% file: sec/2_method_result.tex
\section{Method}
\label{sec:method}
In open‑set recognition (OSR), classifiers are initially trained using only samples from known classes. At test time, they must simultaneously classify known instances and detect and reject previously unseen instances ~\cite{scheirer2013toward}. Earlier OSR methods relied solely on known-class data during training. However, recent research has shown that including a background data set that includes samples from different classes other than known target classes can substantially improve performance~\cite{dhamija2018reducing,miller2021class,cevikalp2023from,geng2021recent}.
We have used the “300k Random Images” dataset~\cite{hendrycks2019oe, freeman2008tinyimages} as background samples (which is a small subset of publicly available Tiny ImageNet~\cite{freeman2008tinyimages} dataset) in our proposed loss function to enforce separation between unknown class regions and known class centers.


\begin{figure}[t]
    \centering
    \includegraphics[width=\columnwidth]{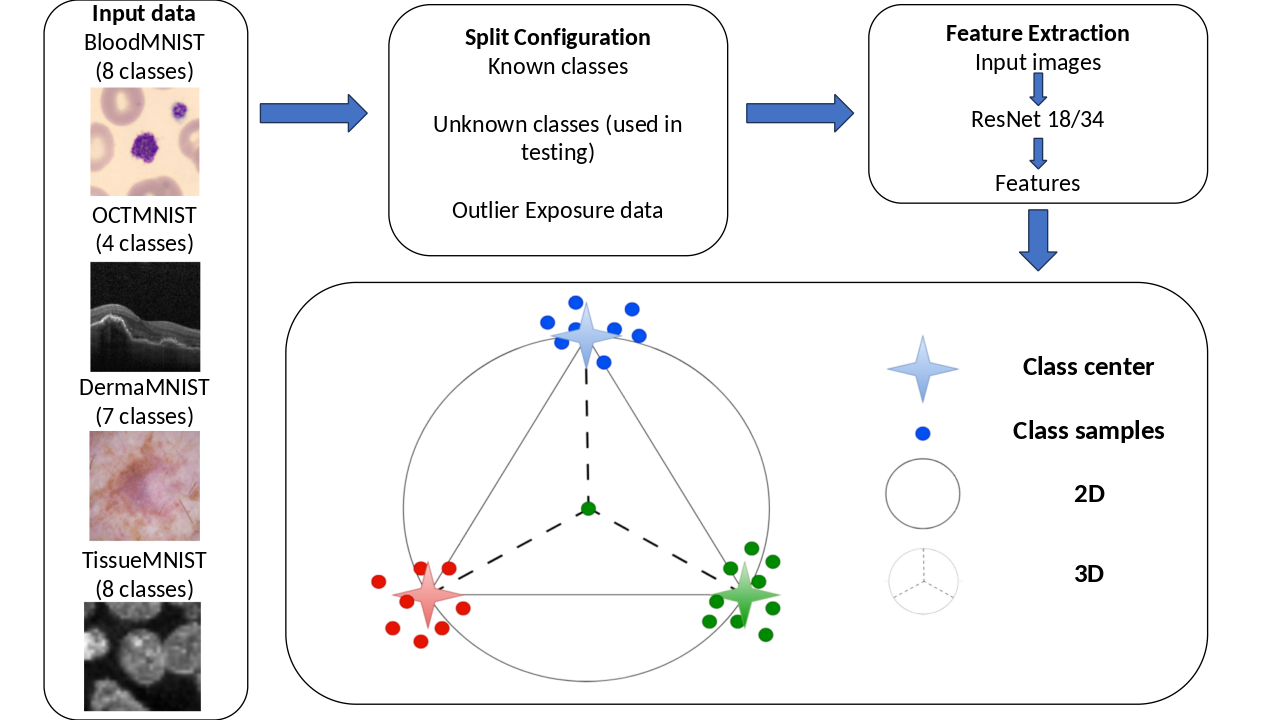}
    \caption{Our method maps extracted features into a geometric space where class centers (stars) are fixed as the vertices of a Simplex ETF, inscribed on the boundary of a hypersphere. This structure ensures that the class centers are maximally separated. During training, the model learns to have class samples (dots) cluster tightly around their corresponding centers. This arrangement provides an intuitive way to measure uncertainty; a hypothetical sample (central green dot) would be equidistant from all class centers, resulting in a maximum uncertainty score of U=1. Conversely, samples mapped confidently near their true class center have an uncertainty score approaching zero ($U \approx 0$)}
    \label{fig:method_overview}
\end{figure}
As mentioned in the introduction section~\ref{sec:intro}, due to neural collapse, class means along with the corresponding classifier weight vectors, align with the vertices of
a simplex equiangular tight frame (ETF). In our method this regular simplex is circumscribed by the hypersphere as shown in figure~\ref{fig:method_overview} (for the case of $2$-D hypersphere circumscribing the equilateral triangle). This $2$-D representation is only for visualization purpose. The radius of this hypersphere is represented by the expand factor which is the hyperparameter in our proposed model.

Let us denote the deep neural feature representations of the training dataset pairs as \((\mathbf{f}_i, y_i)\), for \(i = 1, \dots, n\), where \(\mathbf{f}_i \in \mathbb{R}^d\), \(y_i \in \{1, \dots, C\}\), and \(C\) is the number of known classes. We assume the feature dimensionality \(d\) satisfies \(d \geq C-1\). Under these conditions, the loss function associated with class centers is expressed as:
\begin{equation}
\mathcal{L}_{\text{intra}} = \frac{1}{n} \sum_{i=1}^{n} \left\|\mathbf{f}_i - \mathbf{s}_{y_i}\right\|_2^2
\label{eq:intraclass_loss}
\end{equation}
where \(\mathbf{s}_{y_i}\) represents the target vector (simplex vertex) for class \(y_i\)~\cite{cevikalp2024reaching}. This formulation encourages learned features to align closely with their assigned class centers during training.

Let \(\mathbf{f}^{bg}_k \in \mathbb{R}^d\), \(k = 1,\dots,\mathcal{K}\), denote the deep neural network features of the \(k\)-th background sample. To integrate these background samples into training, an additional loss term is introduced to push background features away from the fixed known-class centers:

\begin{equation}
\mathcal{L}_{\text{o}} =
\sum_{i=1}^{n} \sum_{k=1}^\mathcal{K} 
\max\Bigl(0,\; m + \bigl\|\mathbf{f}_i - \mathbf{s}_{y_i}\bigr\|_2^2 
- \bigl\|\mathbf{f}^{bg}_k - \mathbf{s}_{y_i}\bigr\|_2^2\Bigr)
\end{equation}

where \(m\) denotes a margin hyperparameter.The second loss term $\mathcal{L}_{\text{o}}$ includes a margin constraint, requiring that known class samples remain closer to their respective class centers than background class samples by a minimum margin $m$. 

As in any nearest centroid-based algorithm, the open space between class centers poses the most risk from an open-set classification standpoint~\cite{geng2021recent}, we further included uncertainty aware term $\mathcal{L}_{u}$ in the above mentioned loss function, which penalizes the open space between the class centers.
 This $\mathcal{L}_{u}$ is inspired by ~\cite{cao2021open} and is proportional to the ratio $U$ between the distance to the nearest class center and the average distance to all other class centers. Essentially, $U$ captures the similarity of a sample to the known classes. $U = 1$ signifies that the distance of the latent representation of the test sample from all class centers (vertices of the regular simplex) is the same, which can be interpreted as unclassifiable. This results in imposing a higher penalty on the test sample. If U = 0, the latent representation of the test sample is exactly a class center, which means there is no ambiguity in the classification. Using this loss function on MedMNIST~\cite{medmnistv2} datasets, we observed notable improvements in performance metrics (AUROC, OSCR, and Accuracy) as compared to the state-of-the-art methods.

\begin{equation}
\begin{split}
\mathcal{L}_{\text{u}} &= \frac{1}{n} \sum_{i=1}^{n}
\frac{\min_{j} \left\| f_i - s_j \right\|_2}
{\frac{1}{C-1} \sum_{j \ne j^*} \left\| f_i - s_j \right\|_2}, \\
&\quad j^* \text{: index of the nearest class center}
\end{split}
\label{eq:luncertainty} 
\end{equation}

\begin{equation}
\mathcal{L}_{\text{total}} = \mathcal{L}_{\text{intra}} + \lambda_{\text{o}} \mathcal{L}_{\text{o}} + \lambda_{\text{u}} \mathcal{L}_{\text{u}}
\label{eq:ltotal}
\end{equation}
We evaluated our model using loss function $\mathcal{L}_{\text{total}}$ which is a linear combination of $\mathcal{L}_{\text{intra}}$, $\mathcal{L}_{\text{o}}$, and $\mathcal{L}_{\text{u}}$. We got best results for $\lambda_{\text{o}}$ between $0.1$ and $1.5$, and for $\lambda_{\text{u}}$ between $5$ and $10$.
\section{Experiments and Results}
\label{sec:results}
We first detail our experimental setup, followed by results and discussions. The source code can be accessed at https://github.com/Arnavadi19/UCDSC
\subsection{Experimental Setup}
We start with a brief description about the different datasets used in our study, followed by the implementation details.
\subsubsection{Datasets}

We use MedMNIST v2 datasets~\cite{medmnistv2} and Augmented Skin Conditions Image Dataset \cite{naqvi2023augmented}. In the MedMNIST dataset, all images are resized to $28{\times}28$, and the official training and test splits are used for all four datasets. Each dataset is split into known and unknown classes for each trial. In each trial, images from known classes are used for training and closed-set evaluation. Images belonging to the unknown classes in the test set are used for open-set evaluation.
We used background samples as an auxiliary dataset to discriminate unknown classes from the open space. We considered "300k Random Images" dataset used in \cite{hendrycks2019oe, freeman2008tinyimages} as our background samples.

\textbf{BloodMNIST}~\cite{medmnistv2} is a dataset containing microscopic images of individual normal blood cells.This dataset has been derived from ~\cite{acevedo2020dataset, acevedo2020mendeley}. The images are obtained from individuals without infection or hematologic disorders and are categorized into 8 classes. The dataset contains 17{,}092 color images, center-cropped and resized to 28\,$\times$\,28. 

\textbf{OCTMNIST}~\cite{medmnistv2} is derived from a retinal OCT dataset \cite{kermany2018deep,kermany2018dataset} consisting of 109{,}309 grayscale images categorized into 4 diagnostic classes. The images are center-cropped and resized to 28\,$\times$\,28. 

\textbf{DermaMNIST}~\cite{medmnistv2} is derived from the HAM10000 dataset~\cite{DVN/DBW86T_2018, codella2019skinlesionanalysismelanoma}, a multi-source collection of dermatoscopic images representing 7 common skin diseases. It includes 10{,}015 color images which are resized to 28\,$\times$\,28. 

\textbf{TissueMNIST}~\cite{medmnistv2} originates from the BBBC051 dataset \cite{woloshuk2021in, ljosa2012annotated} from the Broad Bioimage Benchmark Collection. It contains 236{,}386 grayscale images of segmented human kidney cortex cells, labeled into 8 categories. Each 3D image (32\,$\times$\,32\,$\times$\,7) is reduced to 2D via maximum intensity projection along the axial (slice) dimension, and resized to 1\,$\times$\,28\,$\times$\,28 for use. 

\textbf{Augmented Skin Conditions Image Dataset}\cite{naqvi2023augmented} is a collection focused on enhanced images of six common skin conditions. It contains 2,394 images, with 399 images for each condition: acne, carcinoma, eczema, keratosis, milia, and rosacea. The image size is variable, reflecting the characteristics of real-world medical images.

\begin{figure*}[!h] 
    \centering
    
    \begin{subfigure}{\textwidth}
        \centering
        \includegraphics[width=0.95\textwidth]{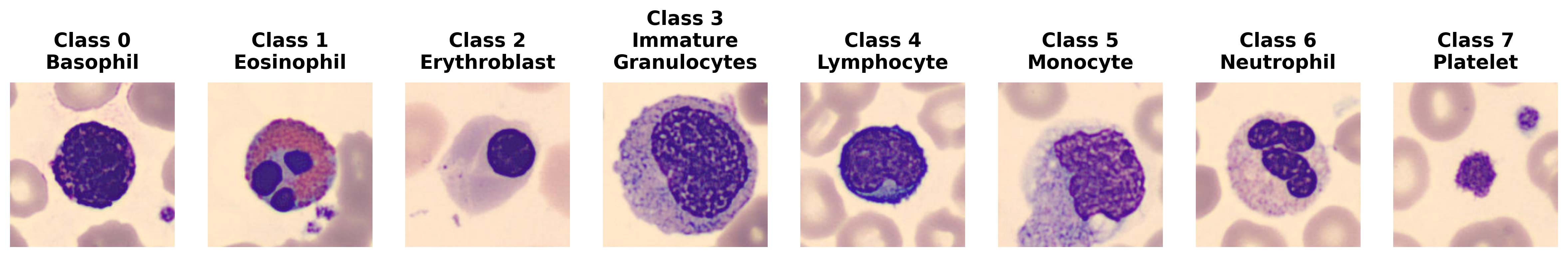}
        \label{fig:bloodmnist-samples}
    \end{subfigure}
    

    \begin{subfigure}{\textwidth}
        \centering
        \includegraphics[width=0.95\textwidth]{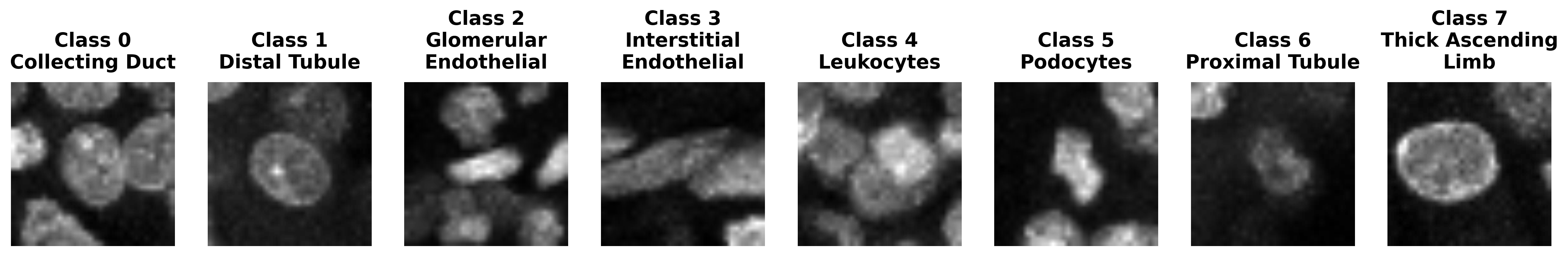}
        \label{fig:tissuemnist-samples}
    \end{subfigure}
    

    \begin{subfigure}{\textwidth}
        \centering
        \includegraphics[width=0.95\textwidth]{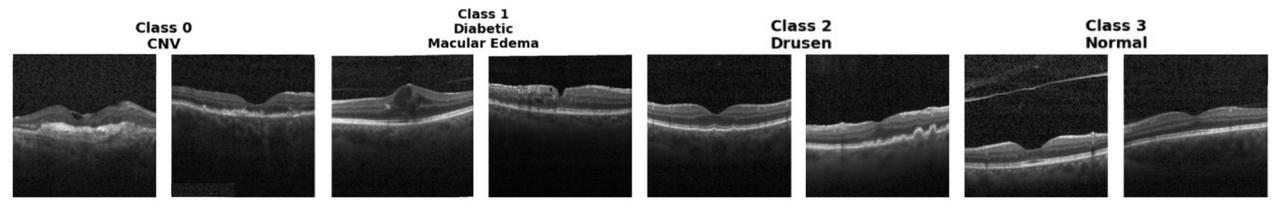}
        \label{fig:octmnist-samples}
    \end{subfigure}
    
    \begin{subfigure}{\textwidth}
        \centering
        \includegraphics[width=0.83\textwidth]{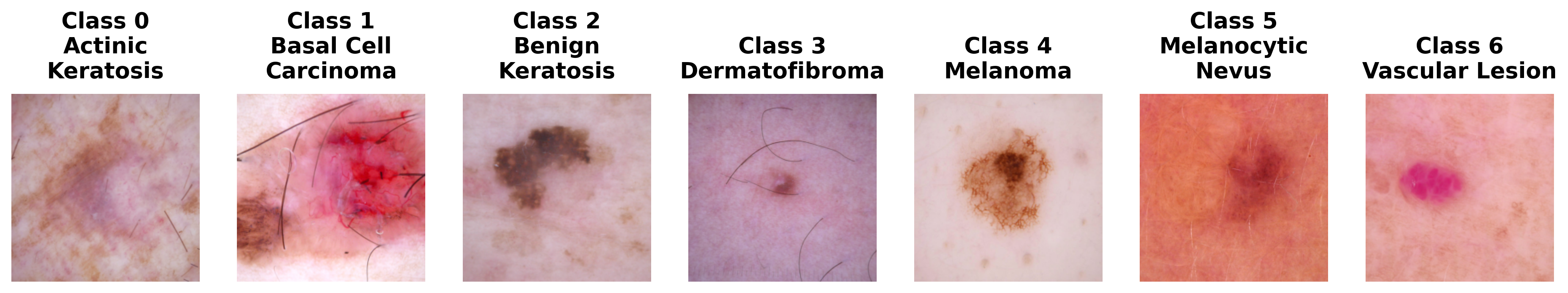}
        \label{fig:dermamnist-samples}
    \end{subfigure}

    \begin{subfigure}{\textwidth}
        \centering
        \includegraphics[width=0.75\textwidth]{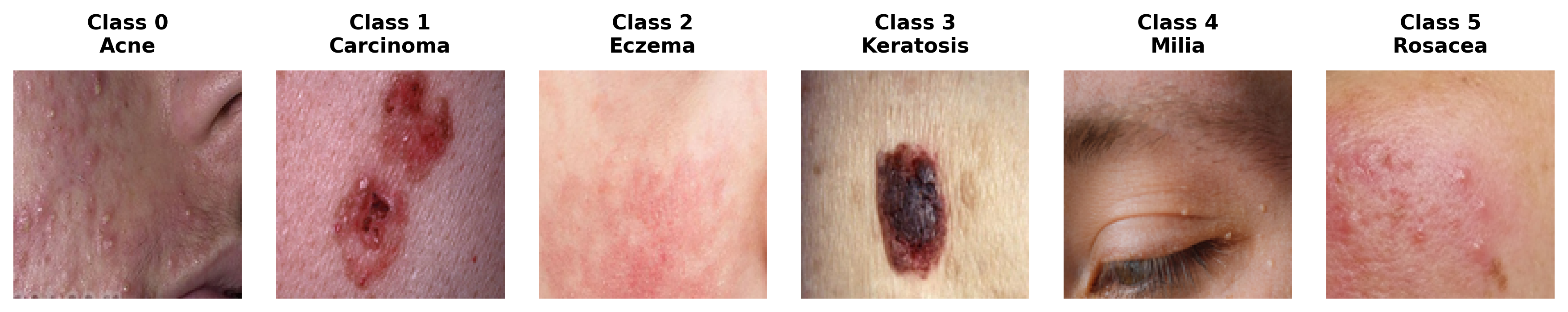}
        \label{fig:asc-samples}
    \end{subfigure}


    \caption{Sample images from the five datasets (a) BloodMNIST (b) TissueMNIST (c) OCTMNIST (d) DermaMNIST (e) Augmented Skin Conditions}
    \label{fig:all-dataset-samples}
\end{figure*}

\subsubsection{Implementation Details}

\paragraph{Metrics} To evaluate performance in both closed and open set scenarios, we adopt three key metrics. \textbf{Accuracy (ACC)} is used to validate closed-set classification performance, computed as the fraction of correctly classified samples among all test samples. \textbf{Area Under the Receiver Operating Characteristic curve (AUROC)}~\cite{FAWCETT2006861} is a threshold-independent metric that assesses open set detection capability by measuring how well the model distinguishes between known and unknown samples. It reflects the probability that a randomly chosen known sample is assigned a higher confidence than a randomly chosen unknown sample. However, AUROC does not account for the correct classification of the known classes. To address this limitation, we additionally use the \textbf{Open Set Classification Rate (OSCR)}~\cite{dhamija2018reducing}, which jointly considers open set recognition and closed set classification performance. OSCR curve plots the Correct Classification Rate (CCR) against the False Positive Rate (FPR) over varying thresholds, providing a comprehensive measure of a model’s ability to classify known samples correctly while rejecting unknowns. OSCR score is the area under the OSCR curve. A higher OSCR score indicates superior open set classification performance.

\begin{figure*}[!h]
    \centering
    \includegraphics[width=0.8\textwidth]{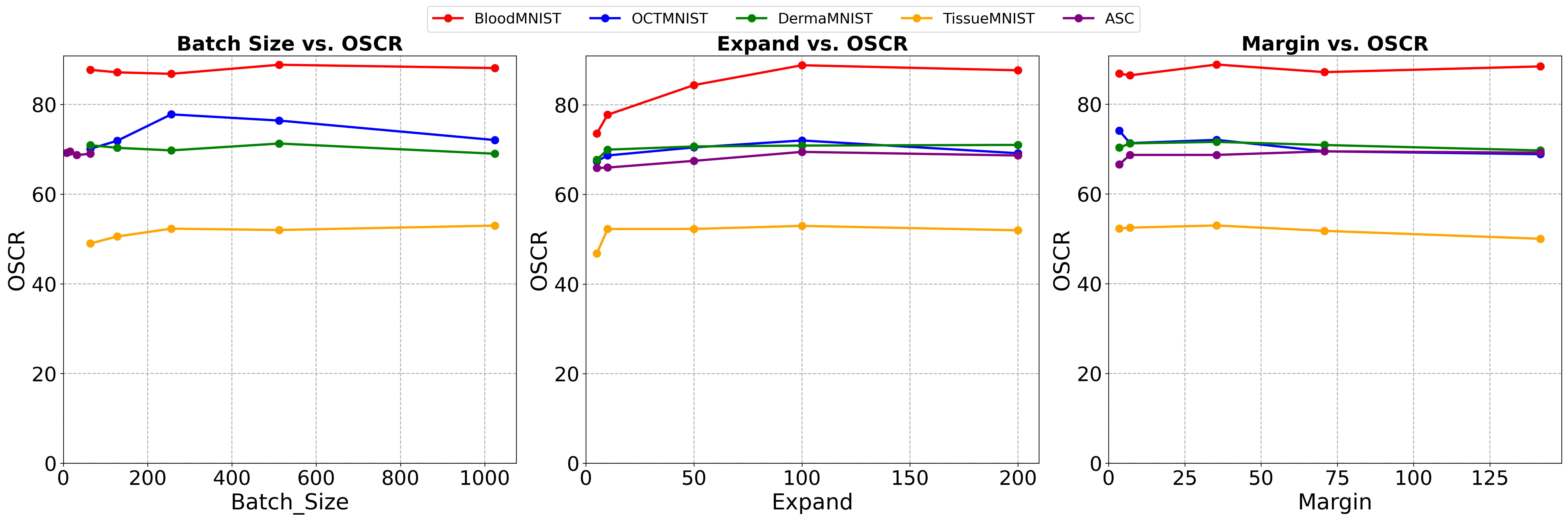}
    \caption{Hyperparameter tuning results for all 5 datasets. These plots illustrate the effect of varying Batch Size, Expand factor and Margin on OSCR metric.}
    \label{fig:hyperparameter-tuning-oscr}
\end{figure*}

\begin{figure*}[!h]
    \centering
    
    \begin{subfigure}{0.19\textwidth}
        \centering
        \includegraphics[width=\linewidth]{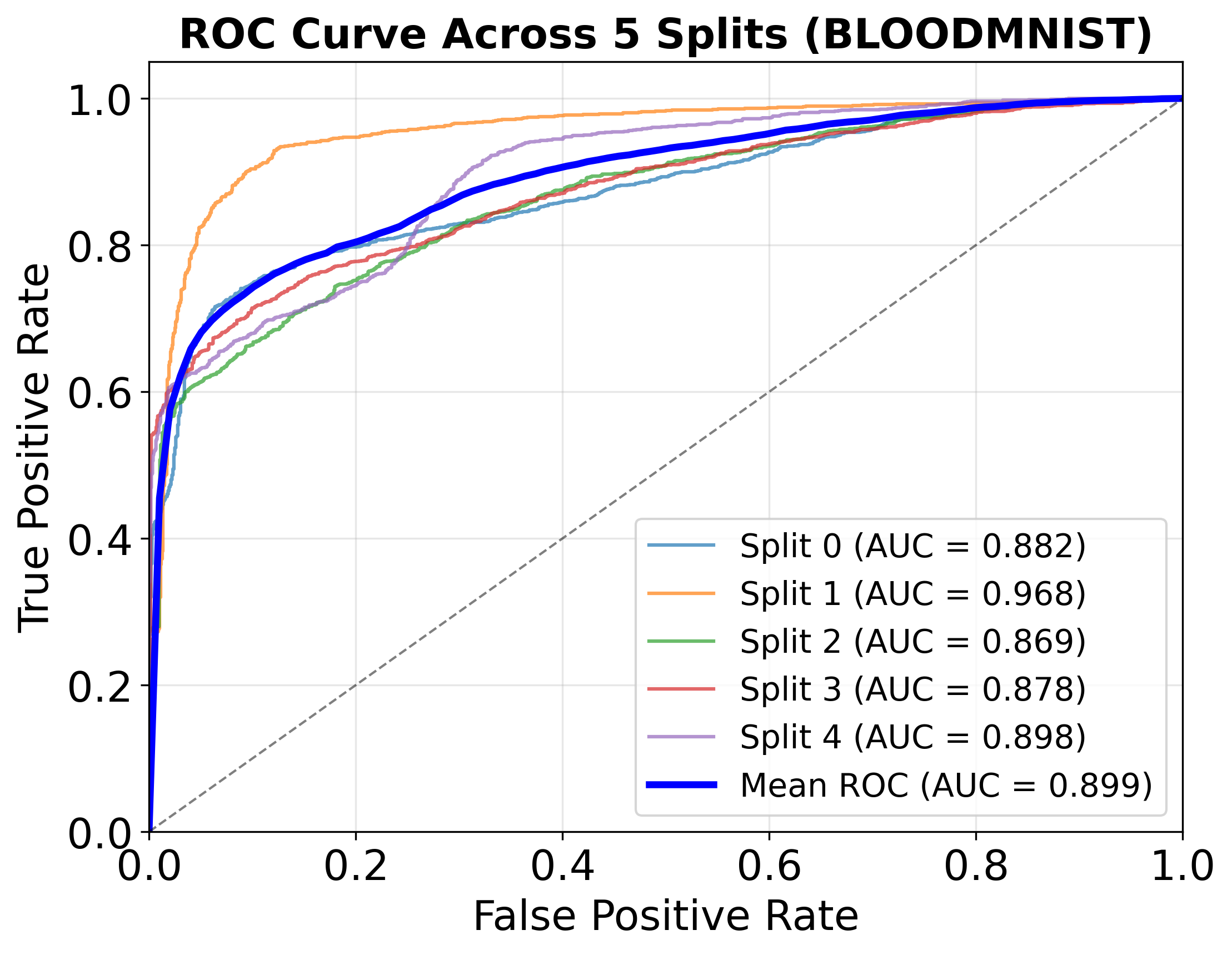}
    \end{subfigure}\hfill
    \begin{subfigure}{0.19\textwidth}
        \centering
        \includegraphics[width=\linewidth]{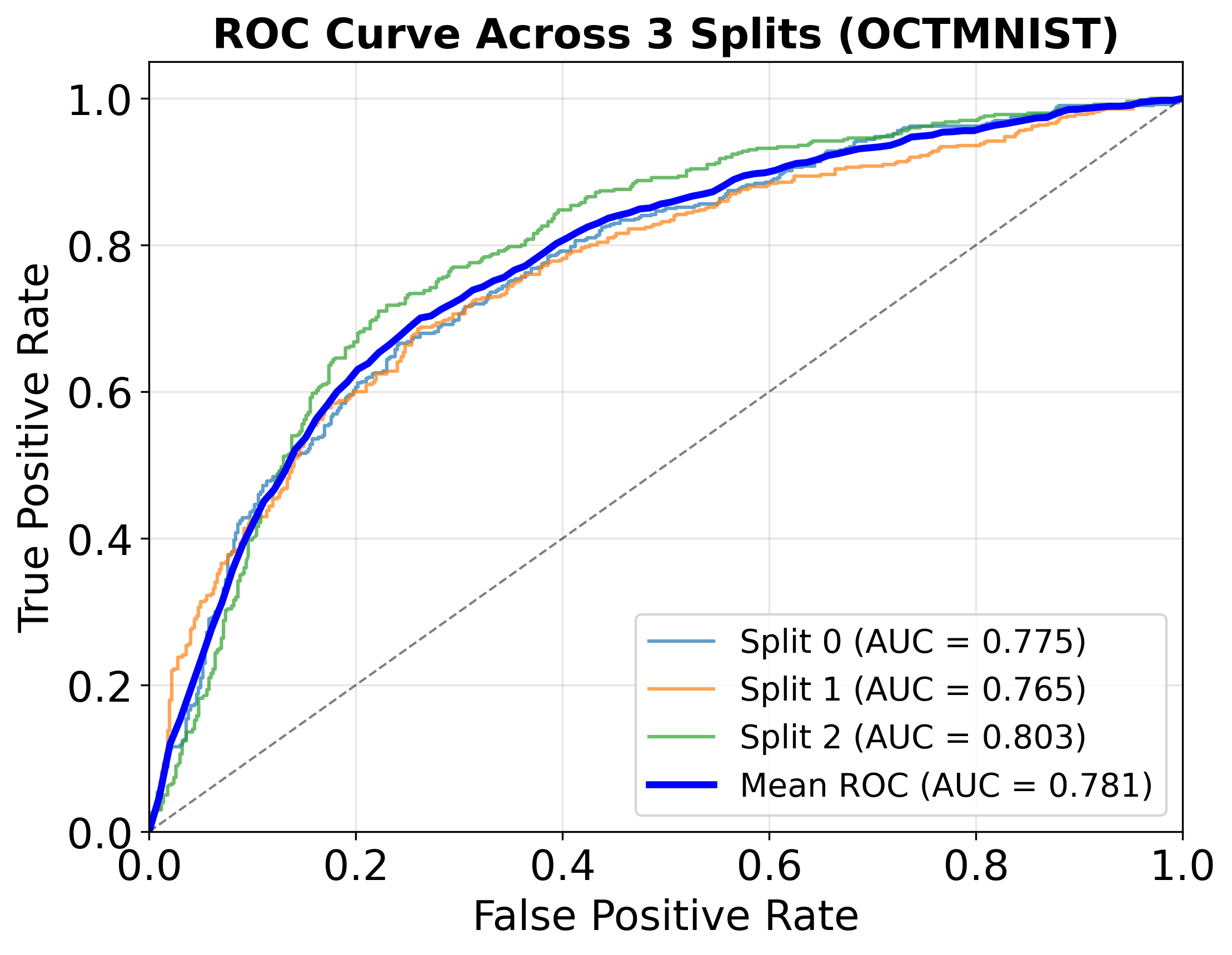}
    \end{subfigure}\hfill
    \begin{subfigure}{0.19\textwidth}
        \centering
        \includegraphics[width=\linewidth]{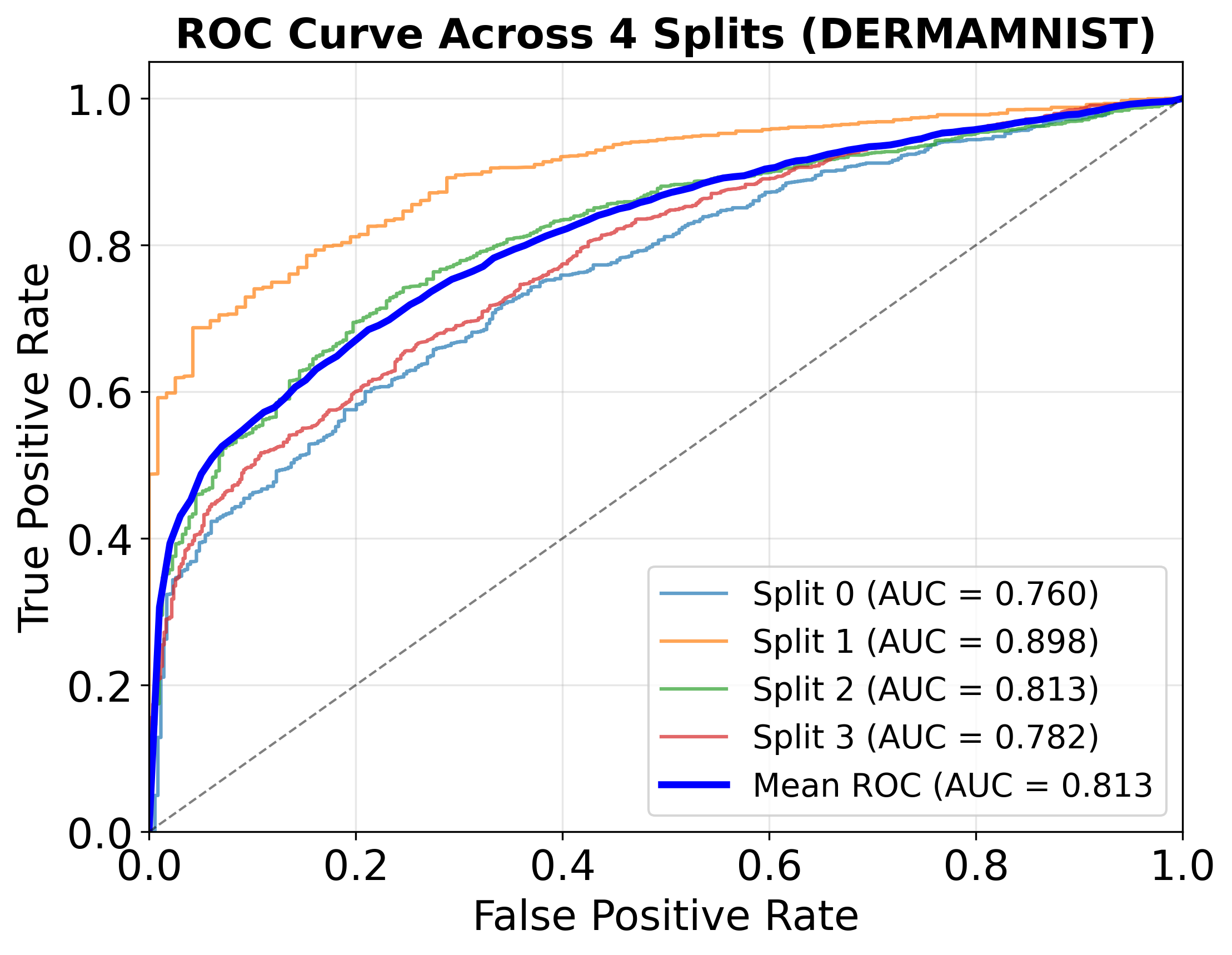}
    \end{subfigure}\hfill
    \begin{subfigure}{0.19\textwidth}
        \centering
        \includegraphics[width=\linewidth]{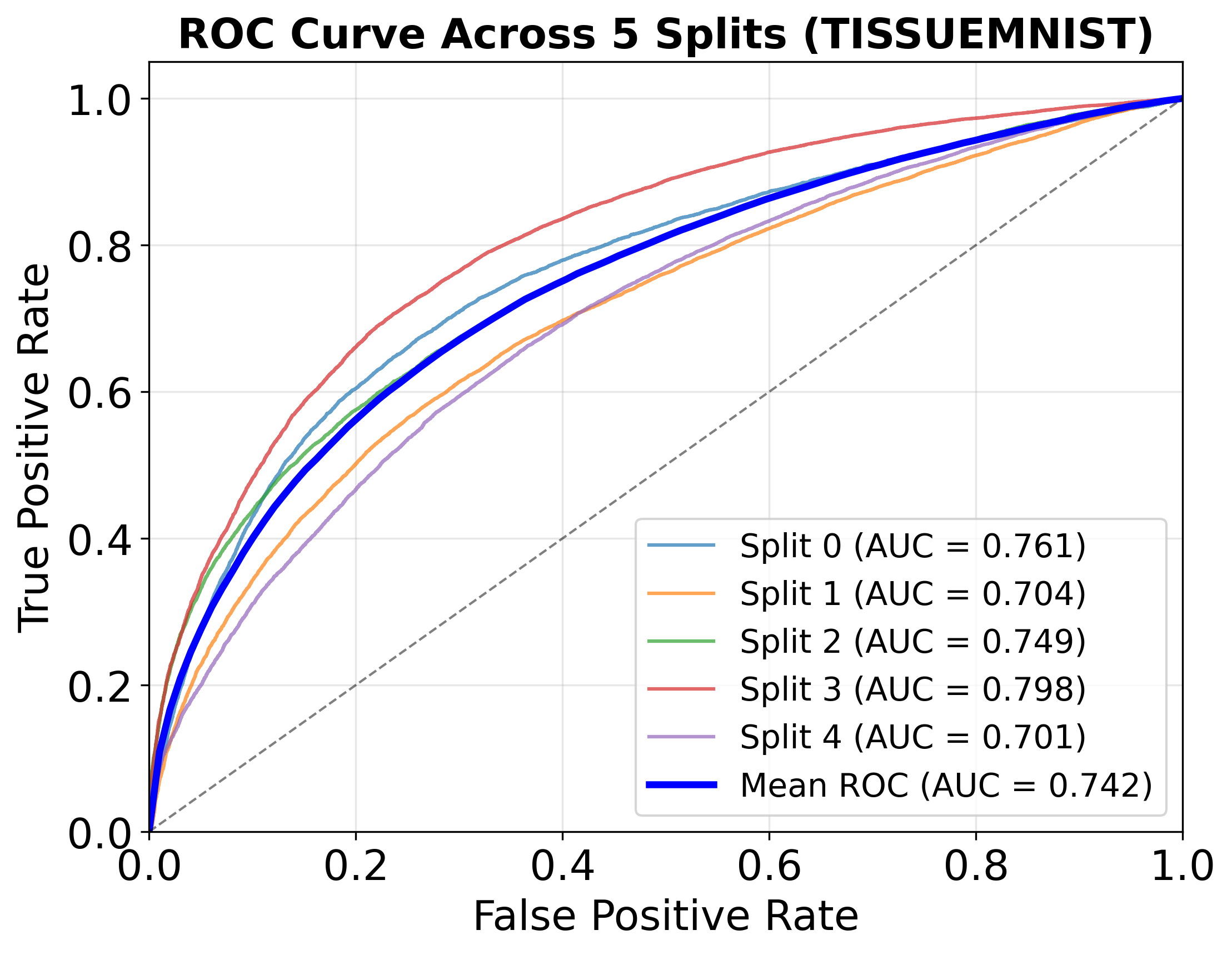}
    \end{subfigure}\hfill
    \begin{subfigure}{0.19\textwidth}
        \centering
        \includegraphics[width=\linewidth]{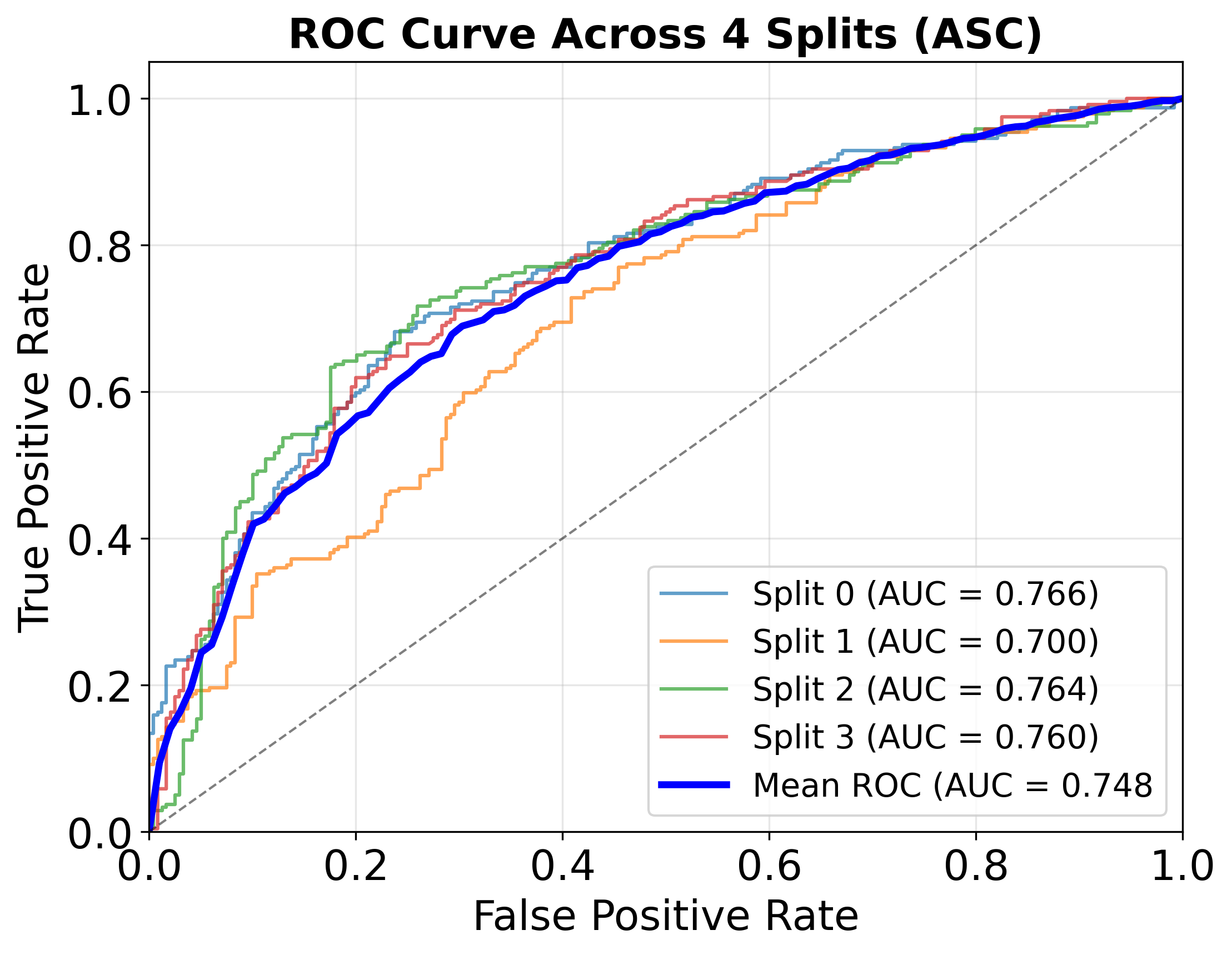}
    \end{subfigure}
    
    \caption{Receiver Operating Characteristic (ROC) curves for five datasets.}
    \label{fig:all-roc-curves}
\end{figure*}

We compared our methods with 6 state-of-the-art methods GCPL~\cite{8578464},  RPL~\cite{10.1007/978-3-030-58580-8_30}, ARPL+CS~\cite{9521769}, DIAS~\cite{moon2022difficulty}, OMCL~\cite{liu2023omcl} and DSC~\cite{cevikalp2024reaching} on BloodMNIST, OCTMNIST, DermaMNIST, and TissueMNIST\cite{10.1007/978-3-030-58580-8_30} datasets  shown in table~\ref{tab:results01}. OMCL~\cite{liu2023omcl} reported results on BloodMNIST and OCTMNIST only. For other datasets, results are unavailable due to the absence of released code and are indicated with “–” in table~\ref{tab:results01}. Best and second-best performances are highlighted in bold-underlined and bold respectively.
Images of size $28\times 28$ were fed as input to the model. Resnet18 and Resnet34 were deployed as backbone networks. Other than the batch size, learning rate and margin hyperparameters, we also tuned on expand factor which is the radius of the hypersphere circumscribing the regular simplex.
Below are the dataset-wise experiment details:

\textbf{BloodMNIST:}We trained the model with batchsize $512$, epochs $400$ and learning rate $0.01$.
Best values of expand factor and margin were recorded as 100 and 38 respectively. We performed $K=5$ trials on randomly chosen $4$ classes as known and rest $4$ as unknown and reported mean value of all three metrics (AUROC, OSCR and Accuracy). It can be clearly observed in table ~\ref{tab:results_comparison_standard} that Uncertainty Aware Deep Simplex (UCDSC) Classifier outperforms the other methods on AUROC and OSCR while it produced at par result in terms of accuracy.

\textbf{OCTMNIST:}We trained the model with batchsize $256$, epochs $400$ and learning rate $0.001$. We performed $K=3$ trials on randomly chosen $2$ classes as known and rest $2$ as unknown and reported mean value of all three metrics (AUROC, OSCR and Accuracy). It can be observed in table~\ref{tab:results_comparison_standard} that UCDSC produced results at par with OMCL while outperforming the other methods. Grayscale OCT images contain fine structural details of retinal layers which might be difficult to be captured through reciprocal points of the prototypes and adversarial based learning. This explains the lower performance of GCPL, RPL, ARPL + CS and DIAS methods. 

\textbf{DermaMNIST:}We trained the model with batchsize $512$, epochs $400$ and learning rate $0.001$. We performed $K=4$ trials on randomly chosen $4$ classes as known and rest $3$ as unknown and reported mean value of all three metrics (AUROC, OSCR and Accuracy). It can be observed in table~\ref{tab:results_comparison_standard} that Uncertainty Aware Deep Simplex Classifier significantly outperformed the other methods on AUROC. One may also note that the DermaMNIST dataset size is relatively small ($10,015$ color images) compared to other $3$ datasets due to which the synthetic data generation might not have been helpful in performance improvement as observed while comparing ARPL + CS, DIAS performance with GCPL and RPL.

\textbf{TissueMNIST:}We trained the model with batchsize $1024$, epochs $300$ and learning rate $0.001$. We performed $K=4$ trials on randomly chosen $4$ classes as known and rest $4$ as unknown and reported mean value of all three metrics (AUROC, OSCR and Accuracy). It can be observed in table~\ref{tab:results_comparison_standard} that Uncertainty Aware Deep Simplex Classifier significantly outperformed the other methods on AUROC. This signify that the unknown class rejection is better executed through our method.

\textbf{Augmented Skin Conditions Image Dataset:} We trained the model with batchsize $16$, epochs $200$ and learning rate $0.01$. We performed $K=4$ trials on randomly chosen $3$ classes as known and rest $3$ as unknown and reported mean value of all three metrics (AUROC, OSCR and Accuracy).

 All models are implemented in PyTorch and trained using the RMSProp optimizer (weight-decay/$l2$-penalty $10^{-3}$, $l1$-penalty $10^{-3}$ alpha $0.95$, eps $10^{-6}$, momentum $0.9$).The corresponding performance on trials is shown using ROC curves in figure~\ref{fig:all-roc-curves}.

\begin{table*}[!h]
\renewcommand{\arraystretch}{1.3}
\centering
\caption{Comparison of different methods across five datasets based on AUROC, OSCR, and ACC metrics. $k$ = number of open-set trials (known \& unknown classes are chosen "randomly" in each trial. Refer Supplementary Material for further details.)}
\label{tab:results_comparison_standard}

\footnotesize 
\setlength{\tabcolsep}{4pt} 
\scalebox{0.9}{ 

\begin{tabular}{|l|c|c|c|c|c|c|c|c|c|c|c|c|c|c|c|}
\hline
\multirow{2}{*}{Methods} & \multicolumn{3}{c|}{BloodMNIST, K=5} & \multicolumn{3}{c|}{OCTMNIST, K=3} & \multicolumn{3}{c|}{DermaMNIST, K=4} & \multicolumn{3}{c|}{TissueMNIST, K=5} & \multicolumn{3}{c|}{\begin{tabular}{@{}c@{}}Augmented Skin \\ Conditions, K=4\end{tabular}} \\ \cline{2-16} 
 & AUROC & OSCR & ACC & AUROC & OSCR & ACC & AUROC & OSCR & ACC & AUROC & OSCR & ACC & AUROC & OSCR & ACC \\ \hline
GCPL~\cite{8578464} & 85.5 & 85 & 98.1 & 65.5 & 64.2 & 94.8 & 70.37 & 62.53 & 81.78 & 44.508 & 25.226 & 48.736 & 54.88 & 34.6 & 52.24 \\ \hline
RPL~\cite{10.1007/978-3-030-58580-8_30} & 86.8 & 86.3 & 98 & 65.9 & 64.2 & 93.7 & 69.93 & 61.76 & 80.785 & 50.93 & 30.14 & 47.75 & 55.61 & 37.55 & 57.86 \\ \hline
ARPL+CS~\cite{9521769} & 87.6 & 87.1 & \textbf{\underline{98.5}} & 77.7 & 75.8 & 95.9 & 73.28 & 67.15 & \textbf{\underline{86.60}} & 68.77 & \textbf{58.59} & \textbf{\underline{79.99}} & 58.16 & 39.27 & 60.39 \\ \hline
DIAS~\cite{moon2022difficulty} & 86.3 & 85.7 & \textbf{98.4} & 74.1 & 72.5 & 96 & 69.7 & \textbf{\underline{74.56}} & 84.93 & 64.75 & \textbf{\underline{68.98}} & 59.91 & 65.65 & \textbf{\underline{72.73}} & 74.39 \\ \hline
OMCL~\cite{liu2023omcl} & 88.6 & \textbf{88} & 98.3 & \textbf{\underline{78.9}} & \textbf{\underline{77.8}} & \textbf{\underline{96.8}} & -- & -- & -- & -- & -- & -- & -- & -- & -- \\ \hline
DSC~\cite{cevikalp2024reaching} & \textbf{89.06} & \textbf{88} & 97.394 & \textbf{78.25} & 74.06 & 91.93 & \textbf{79.07} & 71.06 & \textbf{85.63} & \textbf{70.50} & 57.09 & \textbf{77.38} & \textbf{70.87} & 63.55 & \textbf{86.31} \\ \hline
UCDSC(\textbf{Ours}) & \textbf{\underline{89.93}} & \textbf{\underline{88.84}} & 97.69 & 78.11 & \textbf{76.56} & \textbf{96.7} & \textbf{\underline{81.33}} & \textbf{71.28} & 83.23 & \textbf{\underline{74.25}} & 52.98 & 67.26 & \textbf{\underline{74.8}} & \textbf{69.51} & \textbf{\underline{89.12}} \\ \hline
\end{tabular}
} 

\label{tab:results01}
\end{table*}
\subsection{Ablation Study}
We consider the effect of varying $\lambda_{\text{o}}$ and $\lambda_{\text{u}}$ on the AUROC, OSCR scores and accuracies on BloodMNIST, Augmented Skin Conditions (ASC) and TissueMNIST datasets. It can be observed from the table \ref{tab:ablation_ASC} that adding the triplet loss term $\mathcal{L}_\text{o}$ to $\mathcal{L}_\text{intra}$ ($\lambda_\text{o} \ne 0$ and $\lambda_\text{u} = 0$) results in $1\%-2\%$ performance improvement on ASC dataset while further including the uncertainty-aware term $\mathcal{L}_\text{u}$ ( $\lambda_\text{u} \ne 0$) results in significant improvement ($> 5\%$) in all the scores. Here $\lambda_\text{o} = 0$ corresponds to the absence of background samples. Adding only the uncertainty-aware term $\mathcal{L}_\text{u}$ ( $\lambda_\text{u} \ne 0$) to $\mathcal{L}_\text{intra}$ results in a significant performance gain in the ASC dataset. Specifically, the results on $\lambda_\text{o}=0,\lambda_\text{u}=0.1$ are very close to the best scores at $\lambda_\text{o}=1,\lambda_\text{u}=2$, which shows the effectiveness of uncertainty aware term $\mathcal{L}_\text{u}$ contribution to the $\mathcal{L}_\text{total}$.

In case of the BloodMNIST dataset, it can be observed from table \ref{tab:ablation_bloodMNIST} that adding the triplet loss term $\mathcal{L}_\text{o}$ to $\mathcal{L}_\text{intra}$ ($\lambda_\text{o} \ne 0$ and $\lambda_\text{u} = 0$) results in a significant performance gain (at $\lambda_\text{o}=1$ and $\lambda_\text{u}=0$) while further including uncertainty aware term $\mathcal{L}_\text{u}$ ($\lambda_\text{u} \ne 0$) improves the performance by $1\%-2\%$ (at $\lambda_\text{o}=1$ and $\lambda_\text{u}=5$). This shows that the effect of adding background samples is more prevalent in BloodMNIST dataset. We also observe that adding only $\mathcal{L}_\text{u}$ to $\mathcal{L}_{intra}$  results in a significant performance gain in OSCR and accuracy. 

It can be observed from the table \ref{tab:ablation_TissueMNIST} that on TissueMNIST dataset our proposed method achieves $\sim 70\%$ AUROC scores when $\lambda_\text{o} \ne 0$ and $\lambda_\text{u} \ne 0$ (which seems to be quite stable). This reflects that the proposed method is better at ranking known vs. unknown samples across all thresholds. We get the best results on $\lambda_\text{o}=0.001$ and $\lambda_\text{o}=5$. We also observe that after adding background samples ($\lambda_\text{o} \ne 0$) AUROC scores improve (from table \ref{tab:ablation_TissueMNIST}) while accuracy doesn't vary significantly. Particularly, for $\lambda_\text{o} \ne 1$ and $\lambda_\text{u} = 0$ AUROC and accuracy scores $ 70.50\%$ and $77.38\%$ were observed. 

We also show the hyperparameter plots with respect to the OSCR metric in figures~\ref{fig:hyperparameter-tuning-oscr}.The OSCR score is widely considered to be the most reliable measure for open-set recognition~\cite{dhamija2018reducing}. It is observed that for expand factor $100$ and batch size $512/256$ the optimal AUROC and OSCR values are obtained for all datasets. However, for the margin hyperparameter, the highest AUROC values are obtained for BloodMNIST, OCTMNIST and TissueMNIST at $35$. We get the best results for $16-64$ batch size on ASC dataset.

\begin{table*}[htbp]
\centering
\caption{Results on BloodMNIST with varying $\lambda_o$ and $\lambda_u$.}
\resizebox{\textwidth}{!}{
\begin{tabular}{|l|c|c|c|c|c|c|c|c|c|c|c|c|c|c|c|c|c|c|}
\hline
$\lambda_o$ & 0 & 0 & 0 & 0 & 0 & 0 & 0.001 & 0.001 & 0.001 & 0.01 & 0.01 & 0.01 & 0.1 & 0.1 & 1 & 1 & \textbf{1} & 1 \\ \hline
$\lambda_u$ & 0 & 0.01 & 0.1 & 2 & 5 & 10 & 0 & 1 & 5 & 0 & 0.01 & 2 & 0 & 1 & 0 & 0.01 & \textbf{5} & 10 \\ \hline
AUROC       & 84.16 & 86.82 & 86.89 & 87.85 & 87.57 & 86.83 & 85.23 & 87.78 & 86.32 & 86.67 & 87.73 & 87.69 & 87.96 & 89.95 & 89.06 & 88.77 & \textbf{89.93} & 88.63 \\ 
OSCR        & 79.51 & 84.94 & 83.60 & 85.67 & 84.78 & 85.36 & 83.23 & 86.38 & 85.14 & 85.29 & 86.62 & 86.62 & 86.37 & 88.00 & 88.00 & 87.62 & \textbf{88.84} & 87.46 \\ 
ACC         & 91.63 & 96.30 & 93.60 & 95.85 & 94.33 & 96.69 & 96.09 & 96.75 & 97.21 & 96.37 & 97.49 & 97.52 & 96.58 & 96.33 & 97.39 & 97.58 & \textbf{97.69} & 97.29 \\ \hline
\end{tabular}}
\label{tab:ablation_bloodMNIST}
\end{table*}

\begin{table*}[!h]
\centering
\caption{Results on ASC with varying $\lambda_o$ and $\lambda_u$.}
\resizebox{\textwidth}{!}{
\begin{tabular}{|l|c|c|c|c|c|c|c|c|c|c|c|c|c|c|c|c|c|c|c|c|}
\hline
$\lambda_o$ & 0 & 0 & 0 & 0 & 0 & 0 & 0 & 0.001 & 0.001 & 0.001 & 0.01 & 0.01 & 0.1 & 0.1 & 0.1 & 0.1 & 1 & \textbf{1} & 1 & 1 \\ \hline
$\lambda_u$ & 0 & 0.01 & 0.1 & 1 & 2 & 5 & 10 & 0 & 0.1 & 5 & 0 & 0.01 & 0 & 0.01 & 0.1 & 5 & 0 & \textbf{2} & 5 & 10 \\ \hline
AUROC       & 69.53 & 68.35 & \underline{73.40} & 74.43 & 73.34 & 73.42 & 72.44 & 70.87 & 72.62 & 73.75 & 70.02 & 73.53 & 70.56 & 71.17 & 71.28 & 75.00 & 69.24 & \textbf{74.80} & 73.43 & 72.65 \\
OSCR        & 62.69 & 60.91 & \underline{68.02} & 67.41 & 66.64 & 62.57 & 65.05 & 63.55 & 65.46 & 67.36 & 62.36 & 67.07 & 61.12 & 63.94 & 66.57 & 68.48 & 61.91 & \textbf{69.51} & 65.90 & 65.11 \\
ACC         & 84.73 & 84.00 & \underline{89.54} & 86.82 & 86.30 & 80.34 & 84.12 & 86.31 & 86.30 & 86.82 & 85.57 & 87.03 & 82.65 & 85.98 & 89.23 & 86.72 & 84.84 & \textbf{89.12} & 85.25 & 85.46 \\ \hline
\end{tabular}}
\label{tab:ablation_ASC}
\end{table*}

%
%

\begin{table*}[!t]
\centering
\caption{Results on TissueMNIST with varying $\lambda_o$ and $\lambda_u$.}
\resizebox{0.75\textwidth}{!}{
\begin{tabular}{|l|c|c|c|c|c|c|c|c|c|c|c|c|c|c|}
\hline
$\lambda_o$ & 0 & 0 & 0 & 0 & \textbf{0.001} & 0.01 & 0.01 & 0.1 & 0.1 & 1 & 1 & 1 & 1 & 1 \\ \hline
$\lambda_u$ & 0.01 & 0.1 & 2 & 5 & \textbf{5} & 2 & 5 & 2 & 5 & 0 & 0.1 & 1 & 2 & 5 \\ \hline
AUROC       & 64.25 & 63.65 & 68.79 & \underline{67.32} & \textbf{74.25} & 71.63 & 70.80 & 70.46 & \underline{67.41} & 70.50 & 69.91 & 70.75 & 69.30 & 69.92 \\
OSCR        & 50.39 & 50.56 & 49.36 & \underline{55.03} & \textbf{52.98} & 53.91 & 52.42 & 52.79 & \underline{54.82} & 57.09 & 51.42 & 53.34 & 54.24 & 50.49 \\
ACC         & 72.92 & 73.42 & 65.34 & \underline{74.72} & \textbf{67.26} & 69.63 & 69.32 & 70.11 & \underline{74.58} & 77.38 & 68.31 & 70.33 & 68.87 & 66.70 \\ \hline
\end{tabular}}
\label{tab:ablation_TissueMNIST}
\end{table*}

%% file: sec/3_conclusion.tex
\section{Conclusion}

The objective of this paper is to ensure accurate classification of known class instances while effectively rejecting those belonging to unknown classes. We observed the results of our proposed method UCDSC on a recent medical image dataset MedMNIST. Our method UCDSC includes an uncertainty-aware loss term to the Deep Simplex Classifier (DSC) loss function which further improves the performance on MedMNIST and Augmented Skin Conditions (ASC) Image Dataset. This study can be useful in the healthcare domain where rare diseases, imaging artifacts, or previously unidentified pathologies frequently arise.
As a part of future work, we also plan to extend our method to scenarios where the number of classes may exceed the inherent feature dimensions, such as cancer datasets. 

